\documentclass[conference]{IEEEtran}
\usepackage[brazilian, english]{babel}
\usepackage[utf8]{inputenc}
\usepackage{verbatim}
\usepackage{blindtext, graphicx}
\usepackage{caption}			
\usepackage{subcaption}			
\usepackage{hyperref}
\usepackage[]{algorithm2e}
\SetKw{KwBy}{by}

\usepackage{hyperref}
\usepackage[numbers]{natbib}

\usepackage{multirow}

\usepackage{amsmath}

\usepackage{pgfplots}
\pgfplotsset{width=8cm, compat=1.9}

\begin{document}
\title{Learning to associate detections for real-time multiple object tracking}

\author{\IEEEauthorblockN{Michel Meneses\IEEEauthorrefmark{1},
Leonardo Matos\IEEEauthorrefmark{2}, Bruno Prado\IEEEauthorrefmark{3},
André de Carvalho\IEEEauthorrefmark{4} and
Hendrik Macedo\IEEEauthorrefmark{5}}
\IEEEauthorblockA{PROCC, Universidade Federal de Sergipe and
\IEEEauthorrefmark{4}ICMC, Universidade de São Paulo\\
Email: \IEEEauthorrefmark{1}michel.meneses@dcomp.ufs.br,
\IEEEauthorrefmark{2}leonardo@ufs.br,
\IEEEauthorrefmark{3}bruno@dcomp.ufs.br,
\IEEEauthorrefmark{4}andre@icmc.usp.br,
\IEEEauthorrefmark{5}hendrik@ufs.br}}

\maketitle

\begin{abstract}
	With the recent advances in the object detection research field, tracking-by-detection has become the leading paradigm adopted by multi-object tracking algorithms. By extracting different features from detected objects, those algorithms can estimate the objects' similarities and association patterns along successive frames. However, since similarity functions applied by tracking algorithms are handcrafted, it is difficult to employ them in new contexts. In this study, it is investigated the use of artificial neural networks to learning a similarity function that can be used among detections. During training, the networks were introduced to correct and incorrect association patterns, sampled from a pedestrian tracking data set. For such, different motion and appearance features combinations have been explored. Finally, a trained network has been inserted into a multiple-object tracking framework, which has been assessed on the MOT Challenge benchmark. Throughout the experiments, the proposed tracker matched the results obtained by state-of-the-art methods, it has run 58\% faster than a recent and similar method, used as baseline.
\end{abstract}


\IEEEpeerreviewmaketitle

\section{Introduction}
\label{intro}
\begin{figure}[!b]
\centering
\includegraphics[width=0.8\linewidth]{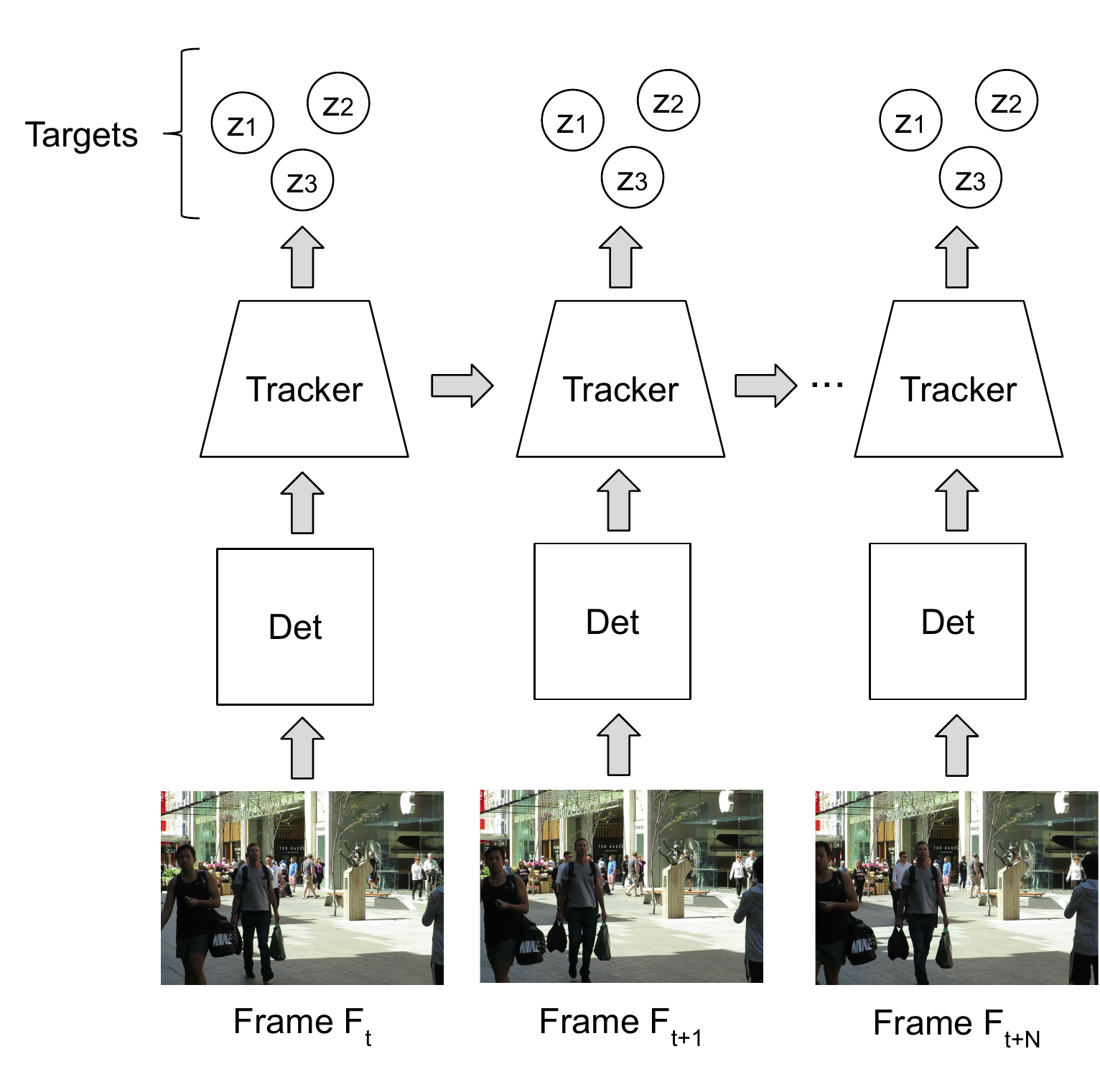}
\caption{Illustration of the tracking-by-detection paradigm. An object detector
(depicted by a square) is applied to the $F_t$ frame at the current $t$ time-step from
a $N$ length video sequence. The tracker uses the detected outputs to identify tracking targets. Next, the same detector is applied to the $F_{t+1}$ frame . The tracker uses the detected outputs to update its set of targets. The procedure is repeated for all $N$ frames.\label{fig:tracking-by-detection}}
\end{figure}

Multiple object tracking (MOT) is a popular topic in Computer Vision due to its wide range of applications (\textit{e.g.}, robotics, autonomous driving vehicles, video surveillance). MOT works by managing the creation and death of new targets, while keeping their identities over a video sequence \cite{Gong2020}. To identify targets with high accuracy, a tracker must solve problems related to illumination changes, camera motion and target occlusions, to name a few \cite{Luo2014}. Given the high accuracy presented by latest object detectors \cite{yolov3, TAO2018}, most of state-of-the-art multiple-object trackers consider the tracking-by-detection paradigm \cite{Yoon2016, Lin2017, Leal-Taixe2017}.

Tracking-by-detection models multiple-object tracking as an association problem between detections extracted from a $N$ length video sequence (\autoref{fig:tracking-by-detection}). Formally, an object detector is applied to the $F_t$ frame obtained at the current $t$ time-step.  Next, the tracker uses the bounding box of each detection to identify and describe the state of its targets. At the next $t+1$ time-step, the same detector is applied to the $F_{t+1}$ frame. The tracker considers the new detections, alongside previously estimated states, to update its set of targets. The procedure is repeated until the detector is applied to all $N$ frames.

One of the main challenges in object tracking is the association problem \cite{Wang2020}. Online tracking-by-detection algorithms usually employ graph-optimization techniques to solve this problem. For such, each set of disjoint $S_t$ vertices represents the detections extracted from the $F_t$ frame, while each edge contains the association cost between a pair of detections from distinctive sets. The set of pairs that minimizes the total association cost can be determined through graph optimization methods, such as network flow \cite{LiZhang2008} and linear programming \cite{Geiger2014} algorithms.

To estimate the association cost between detections, tracking-by-detection algorithms model their targets according to different features, such as appearance and motion. Classical appearance models include the use of pixel templates \cite{Oron2015} and color histograms \cite{Munoz-Salinas2008, Tang2016}. However, models based on convolutional neural networks have shown promising discriminant results, due to their ability to extract deep visual features from detections \cite{Peixia2018}. Other popular motion models include particle filters \cite{Martinez-del-Rincon2011} and Kalman filter \cite{Yu2018}. Additionally, trackers can combine appearance and motion models \cite{Alahi2016}.

Although recent trackers employ several descriptive models to associate detections, their final similarity functions are based on heuristics \cite{Yoon2016, Wojke2017, Bochinski2017}. As a consequence, they present the following drawback: designed functions might be context-domain dependent. Thus, their adaptation to new scenarios is not simple. Besides, being heuristic-based, they are not scalable to handle new descriptor features, which could improve tracking quality.

By looking for patterns in a dataset, Machine Learning (ML) algorithms can discover a similarity function between detections in the context of multiple-object tracking-by-detection \cite{Flach:2012}. Recent works explore ML ability to associate detections; deep learning models have been specifically assessed on this task \cite{Son2017, Sadeghian2017, Alahi2016}. Although they are capable of modeling targets according to several features and estimate their similarity, because of their complex architectures (\textit{i.e.}, CNN and LSTM networks), they are not suitable for online real-time applications.

\begin{figure}[t]
\centering
\includegraphics[width=0.8\linewidth]{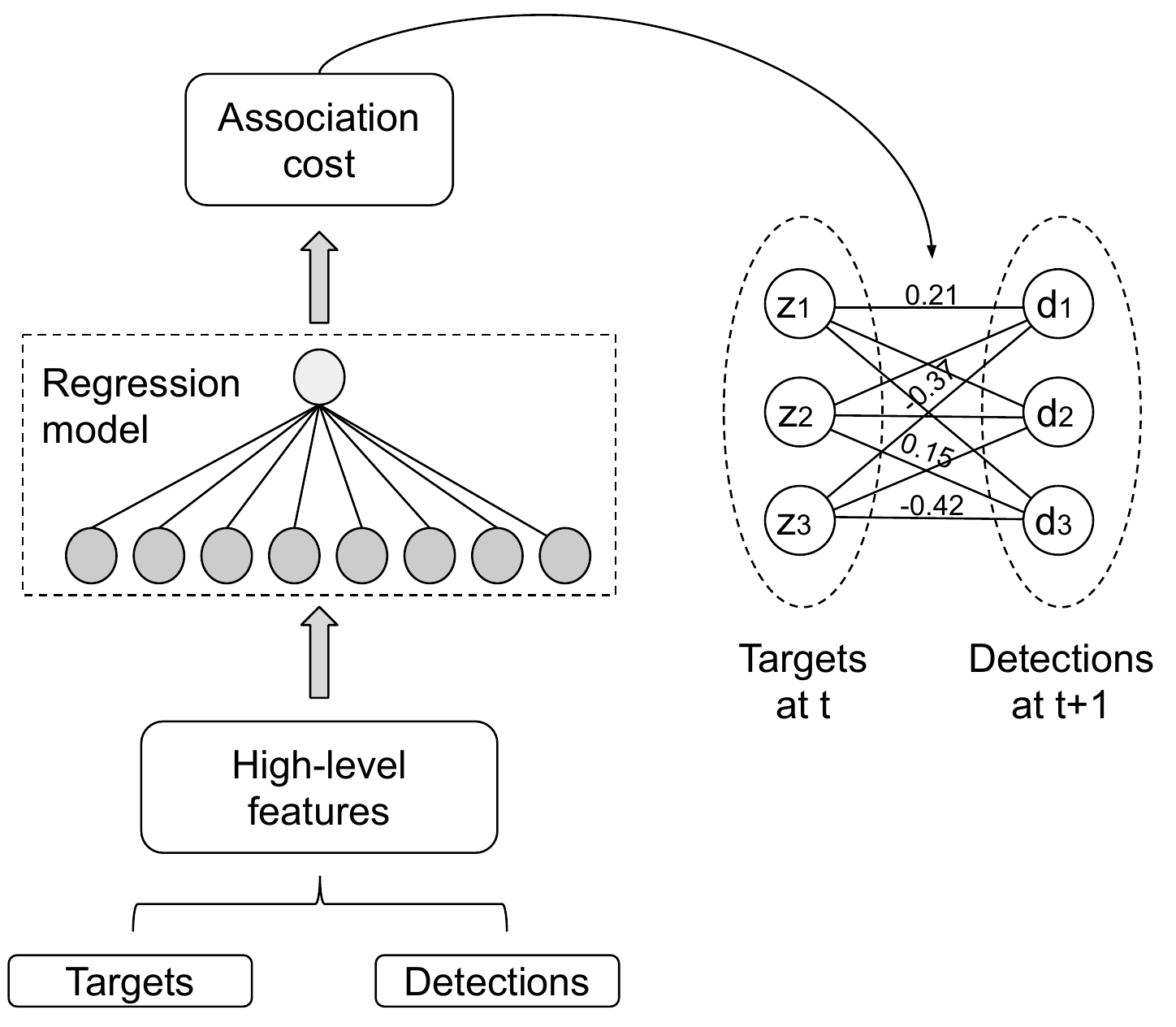}
\caption{Illustration of the proposed method for calculating association costs. Given a set of targets and a set of new detections, a regression model predicts the association cost between each pair of target-detection. This cost is used to construct a bipartite graph, which is solved by a linear programming algorithm.\label{fig:proposed-method}}
\end{figure}

This work investigates a tracking method that uses a simpler, yet feature-scalable and context-adaptable, ML model to estimate the association cost between detections in the tracking-by-detection context (\autoref{fig:proposed-method}). The model receives as input high-level motion and appearance features from the detections. By using ML to learn and adapt the predictive model, the proposed approach is adaptable and scalable. Although any regression algorithm could induce this model, the use of MLP neural networks was defined because of their compact architecture, which is suitable for real-time applications. As discussed later in this paper, the model induced by those networks accomplished low error rates as a regressor, despite of their simplicity. The main benefits of the proposed approach are:

\begin{enumerate}
    \item High adaptability to various scenarios, since a different association cost can be learned for different datasets;
    \item Simple machine learning architecture restricted to calculation of association cost, which is used for online real-time tracking.
\end{enumerate}

The remaining of this paper is organized as follows: Section \ref{sec:mat-meth} describes the main aspects of the proposed method; Section \ref{sec:exp} details its evaluation in a tracking scenario; results are presented in Section \ref{sec:res} and discussed in Section \ref{sec:disc}. Finally, Section \ref{sec:conc} presents the main conclusions.

\section{Proposed method for tracking-by-detection}
\label{sec:mat-meth}

The proposed method, named SmartSORT after the work of \cite{Wojke2017}, adopts an online tracking-by-detection paradigm with frame-by-frame data association, which is described by \autoref{alg:proposed-method}. Its main aspects are presented next.

\begin{algorithm}[t]
 \caption{Proposed SmartSORT tracker. \label{alg:proposed-method}}
 \KwData{$D$ set of new detections at $t+1$ time, $Z$ set of identified targets at $t$ time, $C_\text{max}$ cost threshold, $L_\text{max}$ loss threshold}
 \KwResult{$U$ updated set of identified targets at $t+1$ time}
 \ForEach{$d \in D$}{
 	$d \gets computeAppearanceDescriptor(d)$\;\label{alg:line:appearance-descriptor}
    }
 $U \gets Z$\;
 $C \gets associationCost(Z,D,C_\text{max})$\;\label{alg:line:association-cost}
 $A \gets hungarianMethod(C)$\; \label{alg:line:hungarian-method}
 \ForEach{$(z, d) \in A$}{
   $z \gets updateTarget(z,d)$\; \label{alg:line:target-update}
 }
 \ForEach{$z \in Z$}{
   \If{$z \notin A$}{
     $z \gets incrementLossCounter(z)$\;
     \If{$isTentative(z)$}{
        $U \gets U - \{z\}$\; \label{alg:line:tentative-deletion}
     }
   }
   \If{$getLossCounter(z) > L_\text{max}$}{
     $U \gets U - \{z\}$\; \label{alg:line:target-deletion}
   }
 }
 \ForEach{$d \in D$}{
   \If{$d \notin A$}{
     $z \gets createNewTarget(d)$\; \label{alg:line:target-creation}
     $U \gets U \cup \{z\}$\;
   }
 }
\end{algorithm}

\subsection{Track modeling and handling}

\begin{figure}[t]
\centering
\includegraphics[width=0.9\linewidth]{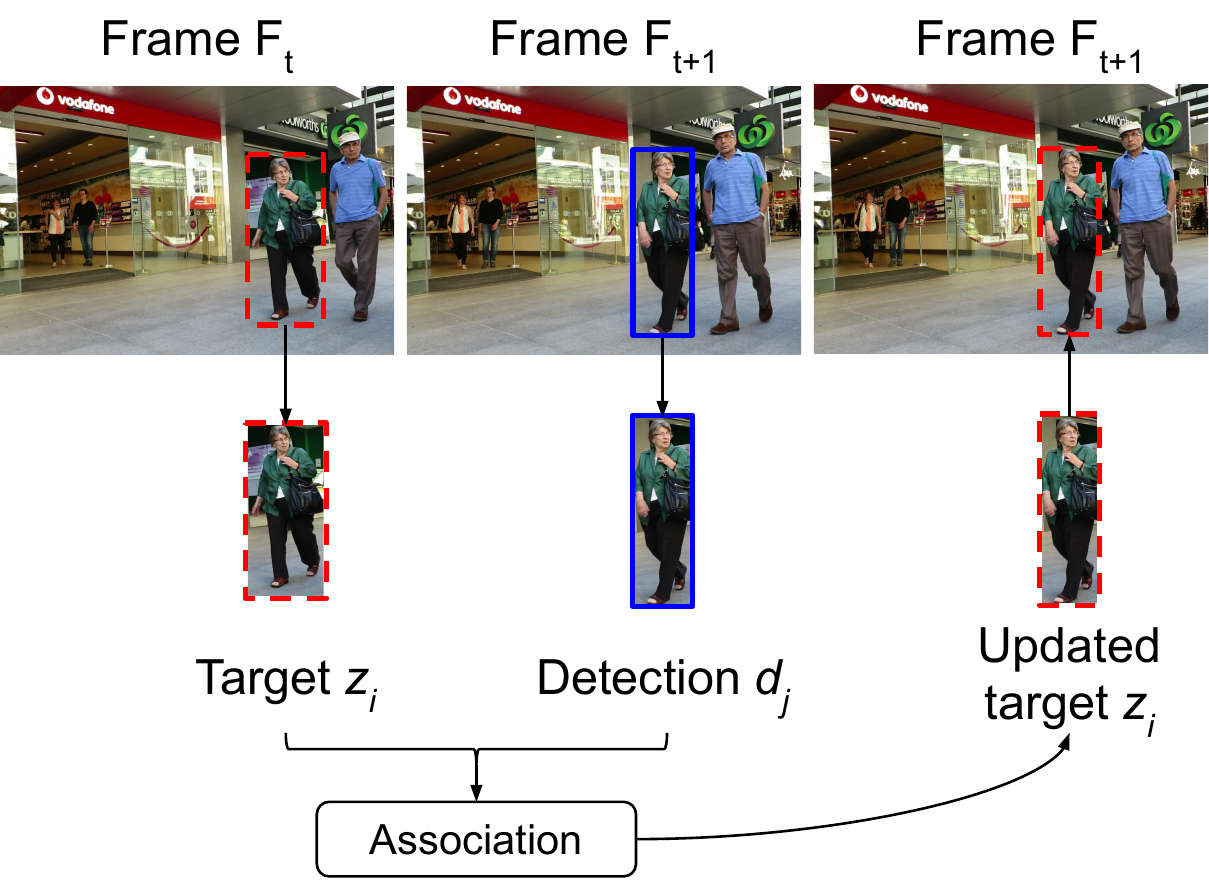}
\caption{Illustration of the association between a $z_i$ target and a $d_j$ detection extracted at $t+1$ time. After their association, $z_i$ incorporates the bounding box and the appearance descriptor of $d_j$. \label{fig:target-update}}
\end{figure}

In this study, it was taken into consideration a single-hypothesis tracking scenario where the state of each $z_i$ target with $i$ id at $t$ time is modeled as:

\begin{equation}
\label{eq:state}
    s_i^t = [u, v, h, r, \boldsymbol{a}]^T
\end{equation}

\noindent where $u$ and $v$ represent, respectively, the horizontal and vertical pixel positions of the center of the target, $h$ stands for the height, $r$ stands for the aspect ratio of its bounding box and finally $\boldsymbol{a}$ denotes its appearance descriptor. The $s_i^t$ target state is updated every time there is an association with a $d_j$ detection (\autoref{alg:line:target-update} of \autoref{alg:proposed-method}). In this case, the target incorporates the detected bounding box, as well as its appearance descriptor (\autoref{fig:target-update}). The former is the output of a CNN framework \cite{Wojke2017}, which computes the deep appearance features of $d_j$ (\autoref{alg:line:appearance-descriptor} of \autoref{alg:proposed-method}). If no association occurs, $z_i$ retains its state.

Our tracker considers the framework proposed by \cite{Bewley2016} for track handling: each $z_i$ target has a $L_i$ loss counter, which is incremented when no associations between a $d_j$ detection and $z_i$ occurs during a tracking iteration, and set to 0 otherwise. If the value of $L_i$ exceeds a given $L_{max}$ threshold, the tracker deletes the $z_i$ target, since it assumes that $z_i$ has permanently left the scene (\autoref{alg:line:target-deletion} of \autoref{alg:proposed-method}). The tracker creates new targets for each detection that cannot be associated (\autoref{alg:line:target-creation} of \autoref{alg:proposed-method}). During their first three frames, these new targets are considered to be tentative. The tracker discards every tentative whose loss counter is incremented (\autoref{alg:line:tentative-deletion} of \autoref{alg:proposed-method}).

\subsection{Data association}

The SmartSORT tracker models the frame-by-frame association between new detections and existing targets as an assignment problem. To compute the association cost between targets and detections, it evaluates motion and appearance information (\textit{i.e.}, their bounding boxes and appearance descriptors). However, unlike related algorithms \cite{POI2016, Wojke2017}, SmartSORT computes this cost using a regression model induced by a machine learning algorithm (\autoref{fig:proposed-method}). In this work, we considered MLP neural networks trained with the Backpropagation algorithm \cite{Cun88}. Thus, given the $\boldsymbol{f}(d_j,z_i)$ feature vector, related to the j-th $d_j$ detection and i-th $z_i$ target, the regression model calculates their $c_{j,i} \in[-1,1]$ association cost. The model calculates this cost for every possible combination of detection-track pair. Subsection \ref{subsec:cost-estimation} details the structure of this model.

Once the regression model has computed every association cost, it optimally solves the assignment problem via the Hungarian method \cite{Kuhn2005} (\autoref{alg:line:hungarian-method} of \autoref{alg:proposed-method}). Additionally, it discards associations whose cost is higher than a $C_{max}$ threshold value, as the tracker admits that they are unfeasible. Since the output space of the regression model is symmetrical, $C_{max}$ has been considered as being 0, so the margin that separates feasible and unfeasible associations can be maximized. It is important to notice that $C_{max}$ is the only hyper-parameter related to the data association step of SmartSORT.

\subsection{Cost estimation regression model}
\label{subsec:cost-estimation}

The main contributions of SmartSORT rely on its cost estimation regression model, which was designed to compute a similarity score between a $z_i$ target and a $d_j$ detection based on their motion and appearance information (\autoref{alg:line:association-cost} of \autoref{alg:proposed-method}). Considering that \autoref{eq:state} expresses the state of $z_i$, the regression model was initially projected to receive as input the $\boldsymbol{f}$ feature vector, defined as:


\begin{equation}
\label{eq:initial_feature}
    \boldsymbol{f}_{d_{j},z_{i}} = [u, v, h, r, \Delta u, \Delta v, \Delta h, \Delta r, \Delta \boldsymbol{a}, \Delta t]
\end{equation}

In \autoref{eq:initial_feature}, $u$, $v$, $h$ and $r$ refer to the bounding box dimensions of $z_i$ target; while $\Delta u$, $\Delta v$, $\Delta h$ and $\Delta r$ represent the normalized differences between the bounding box dimensions of $d_j$ detection and $z_i$; $\Delta \boldsymbol{a}$ is the cosine distance between the $d_j$ and $z_i$ appearance descriptors; and $\Delta t$ measures the number of tracking iterations the $z_i$ target has not being associated with any detection (\textit{i.e.} the value of its $L_i$ loss counter).

The values $u$, $v$, $h$ and $r$ provide to the model information about the absolute position and dimensions of the $z_i$ target, which allows SmartSORT to understand the relation between the motion of $z_i$ and the angle of the camera. On the other hand, $\Delta u$ and $\Delta v$ distances are especially useful for discriminating unfeasible associations between targets, while $\Delta h$ and $\Delta r$ enable SmartSORT to understand their geometry. At the same time, $\Delta \boldsymbol{a}$ distance allows the model to distinguish targets based on their visual cues (\textit{i.e.} their deep appearance features extracted by a CNN framework). Finally, SmartSORT can use $\Delta t$ to understand the temporal dependence of the motion and appearance features of a target, which is particularly useful for occlusion handling.

To increase its capability of motion understanding, it has been decided to expand the input $\boldsymbol{f}$ feature vector to include past positional information. Hence, instead of only considering the distances between positional and visual features of targets and detections at the current time-step, a sliding window strategy has been adopted, where the final $\boldsymbol{g}$ feature vector has the form:


\begin{equation}
    \label{eq:final_feature}
    \begin{split}
    \boldsymbol{g}_{d_{j},z_{i}} = [\boldsymbol{f}_{d_{j},s_{i}^{t}}, \boldsymbol{f}_{s_{i}^{t},s_{i}^{t-1}}
    ,...\ , \boldsymbol{f}_{s_{i}^{t-N-1},s_{i}^{t-N}}]
    \end{split}
\end{equation}

\begin{figure}[t]
\centering
\includegraphics[width=0.9\linewidth]{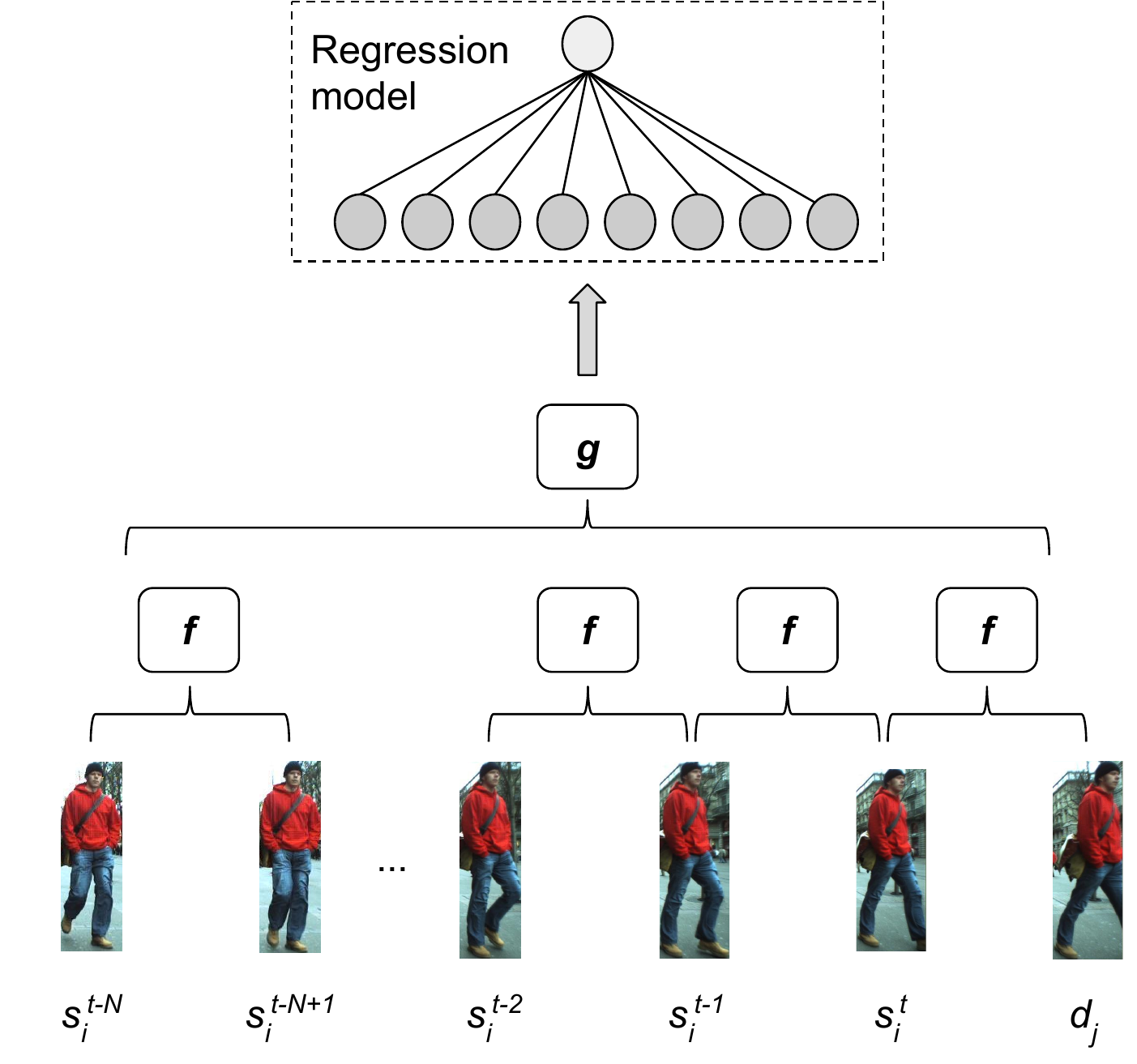}
\caption{Illustration of the sliding window strategy during computation of the cost to associate a $d_j$ detection extracted at $t+1$ and a $z_i$ target. $N$ $\boldsymbol{f}$ feature vectors related to $d_j$ and the states $s_i^k$ of $z_i$ target at $k\in[t-N,t]$ time have been extracted, where $N$ is the length of the temporal window. Afterwards, those vectors have been concatenated to form $\boldsymbol{g}$, which is the input of our regression model at $t+1$ time. \label{fig:sliding-window}}
\end{figure}

In \autoref{eq:final_feature}, $\boldsymbol{f}$ corresponds to the feature vector defined by \autoref{eq:initial_feature}; $d_j$ is the j-th detection extracted at $t+1$ time; $s_i^t$ represents the state of $z_i$ target at $t$ time; and $N$ is the length of the temporal sliding window. This strategy enables the tracker to understand not only the motion behavior of a target but also its temporal appearance variance. \autoref{fig:sliding-window} illustrates our sliding window strategy.

As mentioned before, SmartSORT has been designed to use a model that outputs a $c_{j,i} \in[-1,1]$ similarity score, which represents the association cost between a $d_j$ detection and a $z_i$ target. Although any regression algorithm could induce that model, MLP neural networks trained with the Backpropagation algorithm \cite{Cun88} were employed. The main reason for that was to keep SmartSORT suitable for real-time applications. Also, those networks were able to induce models with low error rates, as discussed in Section \ref{sec:exp}. The model is capable of estimating the association cost between $M$ combinations of targets and detections on a single execution by receiving as input a $M$ feature vectors $B = [\boldsymbol{g}_1, \boldsymbol{g}_2, ..., \boldsymbol{g}_M]^T$ dataset. Algorithm \ref{alg:cost-estimation} describes how the model is applied in the cost estimation step related to the proposed tracking method.

\begin{algorithm}[!b]
 \caption{Association cost estimation. \label{alg:cost-estimation}}
 \KwData{$Z$ set of targets, $D$ set of detections, $C_\text{max}$ cost threshold}
 \KwResult{$C$ set of association costs between targets and detections}
 $B \gets \emptyset$\;
 \ForEach{$z \in Z$}{
    \ForEach{$d \in D$}{
    	$B \gets B + \boldsymbol{g}(d,z)$\;
    	}
    }
 $C \gets regressionModel(B)$\;
 \ForEach{$c \in C$}{
   \If{$c > C_\text{max}$}{
     $C \gets C - c$\;
   }
 }
\end{algorithm}

\section{Experiments}
\label{sec:exp}

A SmartSORT evaluation has three steps: 1) to build an association dataset, 2) to train a MLP neural network and validate its regression model, and 3) to incorporate the model in the method and assess SmartSORT on a tracking benchmark. The following sections describe each of these steps.

\subsection{Association dataset}
A dataset with associations between targets was built by initially sampling positive and negative examples from the seven training sequences in the MOT Challenge Benchmark 2016 \cite{Milan2016}. These sequences have annotations indicating the bounding box and the identity of 517 targets along 5516 frames. For a temporal window of size two, each example has the form:

\begin{equation}
    \label{eq:example_form}
    (s_i^f, s_j^{f+n},\ Y)
\end{equation}

In \autoref{eq:example_form}, $s_i^f$ is the state of $i$ id $z_i$ target at $f$ index frame; $n \in (0, N-f)$ is a random temporal displacement, which is inferior to the difference between the total $N$ frames number, where $z_i$ target is present, and the $f$ index; and $Y$ is the label of the example. Thus, for positive examples, $i=j$ and $Y=-1$, while $i \neq j$ and $Y=1$ for negative examples.

Because this benchmark does not provide detections labeled by their id target, a sampling strategy that considered target-to-target instead of detection-to-target associations was used. After collecting the initial examples, their features have been extracted according to Equations \ref{eq:initial_feature} and \ref{eq:final_feature}. Hence, each final example has the form:

\begin{equation}
    \label{eq:final_example_form}
    (\boldsymbol{g}(z_i, z_j),\ Y)
\end{equation}

In \autoref{eq:final_example_form}, $z_i$ and $z_j$ correspond to targets of $s_i^f$ and $s_j^{f+n}$ states, respectively. Overall, the final dataset has $129\,959$ examples of correct and incorrect associations between targets.

\subsection{Model training}

As part of the experiments with MLP trained by Backpropagation, the hyper-parameters has been tuned using grid-search, including the temporal sliding window length of the input feature vector. The association dataset has been divided into disjoint train and validation partitions. On the former, the sliding window length has been fixed and grid-searches with 3-fold cross-validation have been performed, tuning the number of hidden layers and neurons of the network. After finding the best hyper-parameter values for that specific sliding window length, the MLP network has been trained on the whole training partition and its prediction performance on the validation set has been evaluated.

Concluded these experiments for different lengths of sliding windows, the best model according to its MSE error on the validation set has been selected. The best model had a $8.29\mathrm{e}{-2}$ error. It was an MLP network receiving 40 attribute values, from a sliding window of size 5, and only one 7 neurons hidden layer. To keep the network efficiently trainable \cite{Arora2017UnderstandingDN} and map its output to $[-1,1]$, the activation functions used in its hidden and output layers were, respectively, ReLU and hyperbolic tangent. To induce the final model with Backpropagation Smooth L1 has been employed as loss function, SGD with momentum as the optimizer and a fixed learning rate of $2\mathrm{e}{-3}$. Those parameters were empirically chosen.

\subsection{Tracking evaluation}

The proposed tracker has been evaluated on the testing sequences of the MOT Challenge Benchmark 2016 \cite{Milan2016}. This benchmark assesses the performance of multiple pedestrians trackers on seven different challenging video sequences, each of them presenting various camera setups and lighting conditions. Similarly to \cite{Wojke2017}, our tracker considered as input detections generated by a Faster R-CNN framework \cite{POI2016}. Moreover, similarly to that work, our evaluation test was conducted with $L_{max}=3$ and the detections used as threshold a $0.3$ confidence score. The same CNN framework proposed by \cite{Wojke2017} has also been employed as appearance descriptor for targets and detections.

The benchmark adopts the following metrics to assess the performance of trackers:

\begin{itemize}
    \item Multi-object tracking accuracy (MOTA): the overall accuracy of the tracker in terms of identity switches, false positives and false negatives;
    \item Multi-object tracking precision (MOTP): the precision of the bounding box positions predicted by the tracker;
    \item Mostly tracked (MT): the ratio of ground-truth trajectories that are covered by a track hypothesis for at least 80\% of their respective life span;
    \item Mostly lost (ML): the ratio of ground-truth trajectories that are covered by a track hypothesis for at most 20\% of their respective life span;
    \item Identity switch (ID): the total number of times a ground-truth trajectory was assigned to a different id;
    \item Fragmentation (FM): the total number of times a trajectory was interrupted during tracking.
    \item Runtime: the tracking speed measured in \textit{Hz}, without considering the detection step.
\end{itemize}

Our evaluation test was conducted on an Intel i3-7020U CPU with 4GB of RAM. The appearance features used by SmartSORT were extracted through an Nvidia GeForce GTX 1050 mobile GPU. Since SmartSORT was designed as an improvement of DeepSORT and both methods virtually share the same appearance feature extraction routine, SmartSORT's speed was initially measured without considering the time it takes to extract visual features. To compute its overall runtime speed, we combined its already measured tracking speed with the total time DeepSORT takes to extract appearance features through an Nvidia GeForce GTX 1050 mobile, which is $147s$, as reported by its authors. \cite{Wojke2017}.

\section{Experimental results}
\label{sec:res}

\begin{figure}[!b] 
\centering
\begin{tikzpicture}
\begin{axis}[
    xlabel style={yshift=-20pt},
    xlabel={Frequency [FPS]},
    ylabel style={yshift=20pt},
    ylabel={MOTA [\%]},
    xmin=0, xmax=100,
    ymin=50, ymax=70,
    legend pos=north east,
    ymajorgrids=true,
    grid style=dashed,
    enlargelimits=false,
    clip=false,
]
\addplot[
        scatter,only marks,scatter src=explicit symbolic,
        scatter/classes={
            e={mark=square*,orange},
            f={mark=square*,blue}
        }
    ]
    table[x=x,y=y,meta=label]{
        x y label
        40 61.4 e
        90 60.4 f
    };
    \legend{DeepSORT (baseline), SmartSORT (proposed)}
\draw [orange, opacity=0.5] (0,0) rectangle (40,114);
\draw [blue, opacity=0.5] (0,0) rectangle (90,104.4);
\draw [decorate, decoration={brace, mirror}] 
    ([yshift=-15pt] axis cs:40, 50) -- node[below=3pt] {$\Delta = 50$}
    ([yshift=-15pt] axis cs:90, 50);
\draw [decorate, decoration={brace, mirror}] 
    ([xshift=-15pt] axis cs:0, 61.4) -- node[rotate=90, above=3pt, scale=0.9] {$\Delta = 1$}
    ([xshift=-15pt] axis cs:0, 60.4);
\end{axis}
\end{tikzpicture}
\caption{Comparison between SmartSORT and its baseline performances on the MOT Challenge 2016, assessing tracking accuracy versus runtime frequency. The presented frequencies do not consider the visual feature extraction step performed by both methods. \label{fig:chart-result}}
\end{figure}
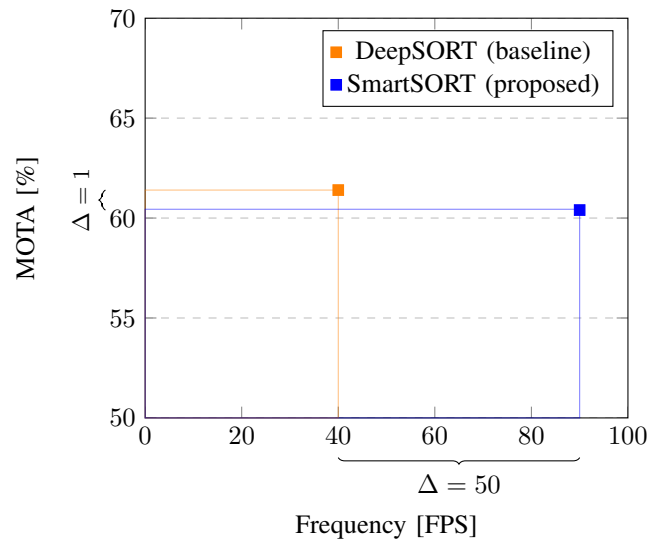

\autoref{fig:chart-result} illustrates a comparison between results obtained by SmartSORT on the MOT Challenge 2016 Benchmark against the performance of its baseline in terms of tracking accuracy and runtime frequency. Since both trackers share the same detection feature extraction routine, their reported speed does not consider the time taken to perform that task.

\autoref{tab:results-mot16} presents the overall results obtained by SmartSORT on that benchmark, as well as the performance of its primary baseline and other online trackers submitted to the same challenge. Among these trackers, there are methods based on motion modeling by Kalman and particle filtering (SORT and EA-PHD-PF) and appearance modeling by deep neural networks (POI, CNNMT and RAN).

\begin{figure}[!b]
	\centering
	\begin{subfigure}[b]{0.45\textwidth}
	\centering
		\includegraphics[width=\linewidth]{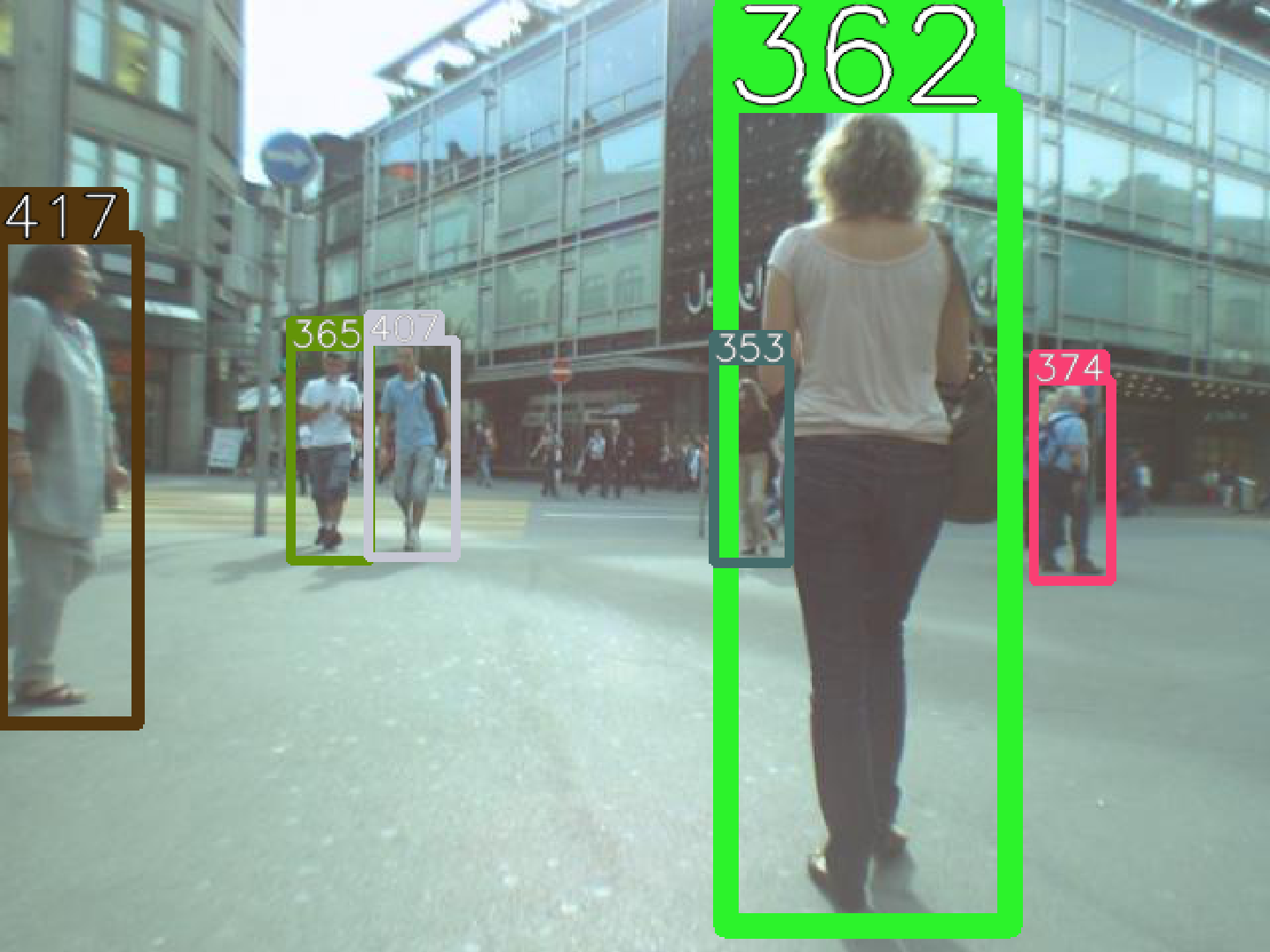}
	\caption{Result on frame 972.\label{fig:qualitative-results-frame-972}} 
	\end{subfigure}
	~
	\begin{subfigure}[b]{0.45\textwidth}
	\centering
		\includegraphics[width=\linewidth]{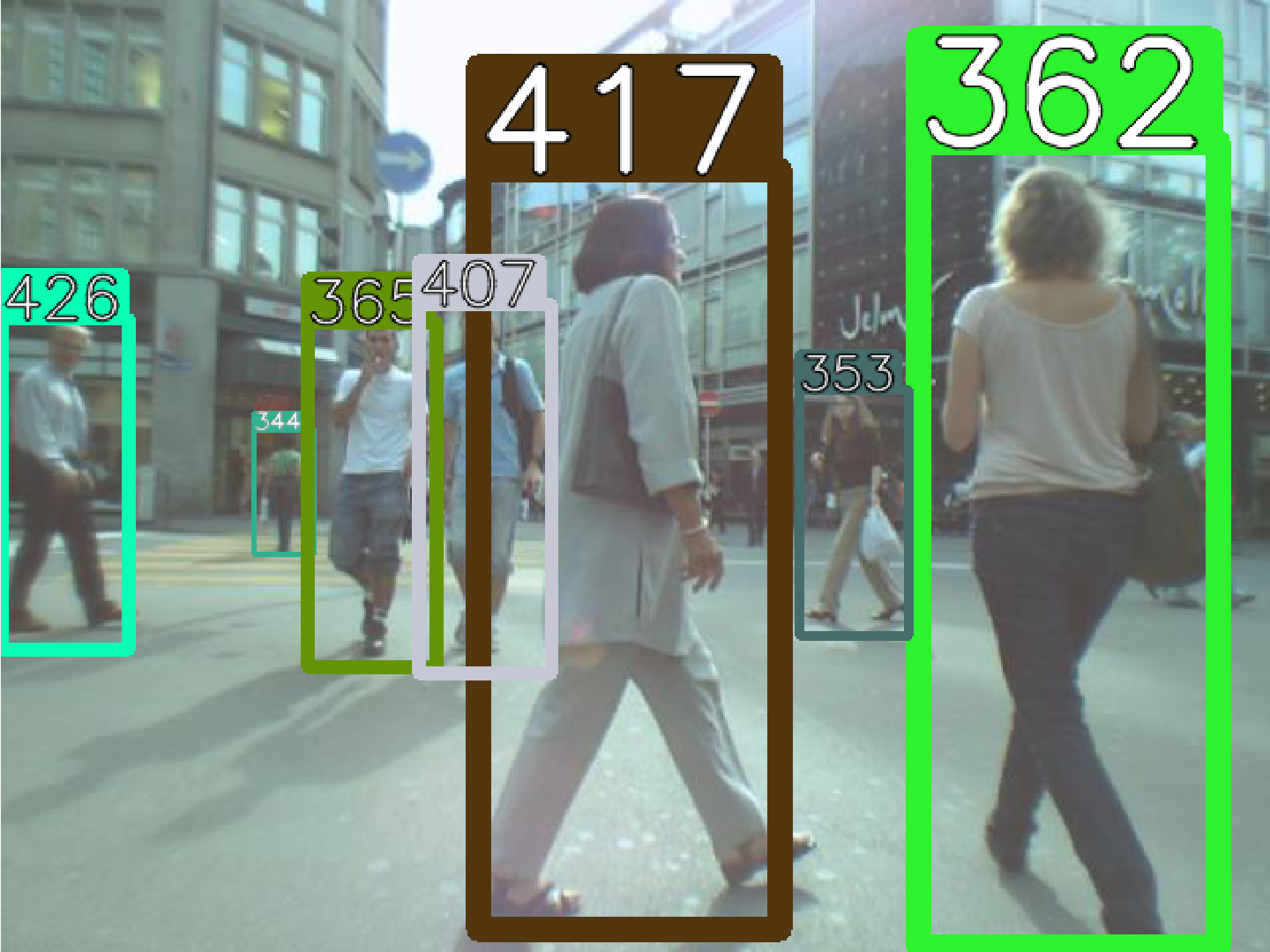}
	\caption{Result on frame 990.\label{fig:qualitative-results-frame-990}}
	\end{subfigure}
	\caption{Qualitative tracking results of the proposed method on the test sequence MOT16-06 from the MOT Challenge 2016.\label{fig:qualitative-results}}
\end{figure}

Finally, \autoref{fig:qualitative-results} illustrates some of the results obtained by SmartSORT when applied to the benchmark.

\begin{table*}[]
\centering
\caption{Online tracking performances on the MOT Challenge 2016. All methods considered private detectors. Also, the presented runtime frequencies do consider the visual feature extraction step performed by those methods.\label{tab:results-mot16}}
\begin{tabular}{lccccccccc} \hline
 &  \textbf{$\uparrow$ MOTA} & \textbf{$\uparrow$MOTP} & \textbf{$\uparrow$MT} & \textbf{$\downarrow$ML} & \textbf{$\downarrow$ID} & \textbf{$\downarrow$FM} & \textbf{$\uparrow$Runtime} \\ \hline
RAN \cite{Fang2017} & 63.0 & 78.8 & 33.9\% & 22.1\% & \textbf{482} & 1251 & 1.6 Hz \\
CNNMT \cite{Nima2018} & 65.2 & 78.4 & 32.4\% & 21.3\% & 946 & 2283 & 11 Hz \\
EA-PHD-PF \cite{Matilla2016O} & 52.5 & 78.8 & 19.0\% & 34.9\% & 910 & 1321 & 12 Hz \\
POI \cite{POI2016} & \textbf{66.1} & 79.5 & \textbf{34.0\%} & 20.8\% & 805 & 3093 & 10 Hz \\
IOU \cite{Bochinski2017} & 57.1 & 77.1 & 23.6\% & 32.9\% & 2167 & 3028 & \textbf{3000 Hz} \\
SORT \cite{Bewley2016} & 59.8 & \textbf{79.6} & 25.4\% & 22.7\% & 1423 & \textbf{1835} & 60 Hz \\
DeepSORT \cite{Wojke2017} & 61.4 & 79.1 & 32.8\% & 18.2\% & 781 & 2008 & 17 Hz \\ \hline
SmartSORT (this paper) & 60.4 & 78.9 & 21.9\% & \textbf{16.1\%} & 1135 & 2230 & 27 Hz \\ \hline
\end{tabular}
\end{table*}

\section{Discussion}
\label{sec:disc}

According to \autoref{fig:chart-result}, SmartSORT, the proposed method, was able to perform its tracking routine at a frequency of 90 FPS against 40 FPS performed by its baseline \cite{Wojke2017}, a 125\% processing speed gain at a 1 percent tracking accuracy cost. This result can be explained by the efficient computation batch performed by its regression model, which calculates the pair-wise association cost between all the currently observed detections and all tracks in a single matrix operation. At the same time, SmartSORT does not perform any filtering computation, as opposed to its baseline. Hence, its tracking management routine has a lower computational cost.

Similar results are obtained when comparing SmartSORT to the other trackers in \autoref{tab:results-mot16}. Even considering the time SmartSORT takes to extract visual features from detections, it runs significantly faster than most of the considered trackers. The only exceptions are IOU and SORT trackers, whose identity switches rates are, at least, 25\% higher than those presented by SmartSORT. On the other hand, SmartSORT runs at least 59\% faster than all the other trackers in \autoref{tab:results-mot16}, which use deep visual features to discriminate detections.

The speed gain of SmartSORT is even higher when compared with the LSTM-based tracker RAN, which extracts features from detections and compute their similarity through a single deep learning framework. This result demonstrates that even though SmartSORT considers a much simpler regression neural network, by presenting high-level deep appearance features and handcrafted motion features as input, it can model a similarity function whose processing time is considerably lower than a more complex network. At the same time, its tracking accuracy was less than 3\% lower than the score obtained by the LSTM-based tracker.

Despite SmartSORT competitive speed and accuracy, its identity switches (ID) and fragmentation scores (FM) were, respectively, $45\%$ and $11\%$ higher than those presented by its baseline. Considering that both methods employ the same appearance feature extractor and the former applies the Kalman filter to encode motion information, this result suggests that the weakest point of SmartSORT is related to its sliding window strategy to predict the trajectory of a target based on its past positions. Thus, that approach may be vulnerable to wrong associations: once SmartSORT switches ids, it contaminates temporal windows with appearance and motion features of distinct targets. This noise may bring instability to its cost estimation, lasting until there is a streak of correct associations. Alternatives to solve this problem include replacing its MLP-based regression model by a time-series specific (\textit{e.g.} vanilla recurrent neural network), while applying compression techniques, such as those based on singular value decomposition (SVD) and adaptive drop weight (ADW) \cite{Prabhavalkar2016, Yang2018}, so its runtime is not impaired.

Nonetheless, by assessing SmartSORT performance on the MOT Challenge 2016 benchmark, it was noticed that it presents high tracking accuracy, at the same time that it runs faster than the other trackers considered in this work. As a result, its overall cost-effectiveness is highly competitive, especially when considering its use in online tracking applications based on embedded systems.

\section{Conclusions}
\label{sec:conc}

In this paper, we propose SmartSORT, a new online tracking-by-detection algorithm that uses a regression model to predict the association cost between detections based on high-level appearance and motion features. Since SmartSORT encodes these features through a temporal sliding window, it can run without need of a filtering algorithm. Results obtained from experimental evaluations of SmartSORT on the MOT Challenge benchmark have shown that its tracking accuracy is competitive with state-of-the-art online trackers, but at a lower computational cost. Therefore, SmartSORT presents highly competitive cost-effectiveness for online real-time tracking applications.

As future research, we intend to experiment the use of shallow machine learning architectures specifically designed to modeling sequential data (\textit{e.g.} vanilla recurrent neural networks) instead of an MLP combined with sliding window. Moreover, in order to keep it suitable for online real-time tracking, we intend to investigate the use of compression techniques while still presenting high-level input features to the model. We also intend to experiment with the use of different training algorithms besides Backpropagation. We believe that those changes will boost the efficiency of training and improve the robustness of SmartSORT to wrong associations, which will increase its overall tracking accuracy.

\noindent\small{\textbf{Funding} This study has been financed in part by the Coordenação de Aperfeiçoamento de Pessoal de Nível Superior - Brasil (CAPES) - Finance Code 001. This research was carried out using the computational resources of the Center for Mathematical Sciences Applied to Industry (CeMEAI) funded by FAPESP (grant 2013/07375-0).}

\bibliographystyle{plain}
\bibliography{referencies}

\end{document}